\documentclass[dvipsnames,format=sigconf,anonymous=false,review=false]{acmart}
\renewcommand\footnotetextcopyrightpermission[1]{} % removes footnote with

\usepackage{listings}

\definecolor{mGreen}{rgb}{0,0.6,0}
\definecolor{mGray}{rgb}{0.5,0.5,0.5}
\definecolor{mPurple}{rgb}{0.58,0,0.82}
\definecolor{backgroundColour}{rgb}{0.95,0.95,0.92}

\definecolor{codegreen}{rgb}{0,0.6,0}
\definecolor{codegray}{rgb}{0.5,0.5,0.5}
\definecolor{codepurple}{rgb}{0.58,0,0.82}
\definecolor{backcolour}{rgb}{0.95,0.95,0.92}

\lstdefinestyle{Pystyle}{
    backgroundcolor=\color{backcolour},   
    commentstyle=\color{codegreen},
    keywordstyle=\color{magenta},
    numberstyle=\tiny\color{codegray},
    stringstyle=\color{codepurple},
    basicstyle=\ttfamily\footnotesize,
    breakatwhitespace=false,         
    breaklines=true,                 
    captionpos=b,                    
    keepspaces=true,                 
    numbers=left,                    
    numbersep=5pt,                  
    showspaces=false,                
    showstringspaces=false,
    showtabs=false,                  
    tabsize=2
}

\lstdefinestyle{CStyle}{
    backgroundcolor=\color{backgroundColour},   
    commentstyle=\color{mGreen},
    keywordstyle=\color{magenta},
    numberstyle=\tiny\color{mGray},
    stringstyle=\color{mPurple},
    basicstyle=\footnotesize,
    breakatwhitespace=false,         
    breaklines=true,                 
    captionpos=b,                    
    keepspaces=true,                 
    numbers=left,                    
    numbersep=5pt,                  
    showspaces=false,                
    showstringspaces=false,
    showtabs=false,                  
    tabsize=2,
    language=C
}

\AtBeginDocument{%
  \providecommand\BibTeX{{%
    \normalfont B\kern-0.5em{\scshape i\kern-0.25em b}\kern-0.8em\TeX}}}

\begin{document}

%%
%% The "title" command has an optional parameter,
%% allowing the author to define a "short title" to be used in page headers.
\title{Initial Steps Towards Tackling High-dimensional Surrogate Modeling for Neuroevolution Using Kriging Partial Least Squares}

%%
%% The "author" command and its associated commands are used to define
%% the authors and their affiliations.
%% Of note is the shared affiliation of the first two authors, and the
%% "authornote" and "authornotemark" commands
%% used to denote shared contribution to the research.
\author{Fergal Stapleton}
%%\authornote{Both authors contributed equally to this research.}

\authornote{Joint first authors}
\email{fergal.stapleton.2020@mumail.ie}
\orcid{0000-0002-5347-1573}
\affiliation{%
  \institution{Hamilton Institute, Maynooth University, \\ Naturally Inspired Computation Res. Group,}
  %%\streetaddress{P.O. Box 1212}
  %%\city{Kildare}
  %%\state{Ohio}
  \country{Ireland}
  %%\postcode{43017-6221}
}
\author{Edgar Galv\'an}
%%\authornotemark[1]
\authornotemark[1]
%%\authornote{Joint first authors}
\email{  edgar.galvan@mu.ie}
\affiliation{%
  \institution{Dept. of CS, Hamilton Institute, IVI, Maynooth University, \\ Naturally Inspired Computation Res. Group,}
  %%\streetaddress{P.O. Box 1212}
  %%\city{Dublin}
  %\state{Kildare}
  \country{Ireland}
  %%\postcode{43017-6221}
}

%%
%% The abstract is a short summary of the work to be presented in the
%% article.
\begin{abstract}
Surrogate-assisted evolutionary algorithms (SAEAs) aim to use efficient computational models with the goal of approximating the fitness function in evolutionary computation systems. This area of research has received significant attention from the specialised research community in different areas, for example, single and many objective optimisation or dynamic and stationary optimisation problems. An emergent and exciting area that has received little attention from the SAEAs community is in neuroevolution. This refers to the use of evolutionary algorithms in the automatic configuration of artificial neural network (ANN) architectures, hyper-parameters and/or the training of ANNs. However, ANNs suffer from two major issues: (a) the use of highly-intense computational power for their correct training, and (b) the highly specialised human expertise required to correctly configure ANNs necessary to get a well-performing network. This work aims to fill this important research gap in SAEAs in neuroevolution by addressing these two issues. We demonstrate how one can use a Kriging Partial Least Squares method in place of the well-known Kriging method, which normally cannot be used in neuroevolution due to the high dimensionality of the data.
\end{abstract}
%%
%% The code below is generated by the tool at http://dl.acm.org/ccs.cfm.
%% Please copy and paste the code instead of the example below.
%%
\begin{CCSXML}
<ccs2012>
<concept>
<concept_id>10010147.10010257.10010293.10011809</concept_id>
<concept_desc>Computing methodologies~Bio-inspired approaches</concept_desc>
<concept_significance>500</concept_significance>
</concept>
<concept>
<concept_id>10010147.10010257.10010293.10010294</concept_id>
<concept_desc>Computing methodologies~Neural networks</concept_desc>
<concept_significance>500</concept_significance>
</concept>
<concept>
<concept_id>10002950.10003648</concept_id>
<concept_desc>Mathematics of computing~Probability and statistics</concept_desc>
<concept_significance>500</concept_significance>
</concept>
</ccs2012>
\end{CCSXML}

\ccsdesc[500]{Computing methodologies~Bio-inspired approaches}
\ccsdesc[500]{Computing methodologies~Neural networks}
\ccsdesc[500]{Mathematics of computing~Probability and statistics}
%%
%% Keywords. The author(s) should pick words that accurately describe
%% the work being presented. Separate the keywords with commas.
\keywords{Neuroevolution, Neural Architecture Search, Surrogate-assisted Evolutionary Algorithms, Kriging, Partial Least Squares}

%% A "teaser" image appears between the author and affiliation
%% information and the body of the document, and typically spans the
%% page.

%%
%% This command processes the author and affiliation and title
%% information and builds the first part of the formatted document.
\maketitle

\section{Introduction}

Artificial Neural Networks (ANNs) are a branch of ML algorithms originally designed to mimic how the human brain works. They have revolutionised the way researchers and society alike tackle complex and challenging problems, for example, they are used extensively in the emerging field of autonomous driving~\cite{neurotraj2022fergal}. However, there are a number of issues associated with using these ANNs algorithms. Firstly, is in the training of these networks which often requires heavy computation resources to train ANNs effectively. Secondly, is the difficulty in designing ANNs’ network architectures \textit{a priori}, often leading practitioners to rely on intuition and random search in the design process. 

The first of these issues, regarding the extensive computational time to train these ANNs, can be addressed using surrogate-assisted evolutionary algorithms~\cite{jin2011surrogate}. The aim of these models is to sufficiently estimate the correct fitness value of a potential solution, while reducing the run time of the overall process, when compared to using the fitness function in the entire set of potential solutions. This requires that the surrogate model is well-informed and has a robust model management strategy. The model may converge to a false optimum~\cite{jin2011surrogate}, otherwise. The second issue, related to finding a correct architecture to have well-performing ANNs, can be addressed using neuroevolution~\cite{Galvn2021NeuroevolutionID}. The latter refers to a branch of evolutionary algorithms (EAs) techniques~\cite{eiben_2015_from} designed to search and evolve ANNs’ hyperparameters, weights and/or architectures.

A major challenge remains, however, at the intersection of the aforementioned two issues, in that it can be difficult to generate input data for a surrogate model based on the architecture information alone. The evolvable structure, or genotypic representation, can often result in mismatches when comparing networks, since network sizes, weights, and parameters may vary. A novel workaround to this limitation is the work by Stork et al.~\cite{pheno_dist_kernel}, which makes use of a phenotypic distance vector that instead relies on the behaviour of the network. Stork et al.,  work demonstrated that neuroevolutionary approaches need not be tied directly to the genotypic architecture. It is fair to say, however, that one limitation of their work is the costly implementation of the Kriging approach used. For example, the computational cost increases by $O(m^3)$ where $m$ is the surrogate model sample size. In order to counteract this issue they subset the surrogate model samples from a larger archive. Their results were promising for subset sizes of 25 to 200. However, it is likely for more complex networks, with larger search spaces, the surrogate may require larger sample sizes in order to be well-informed.

The main contributions of this work are, firstly, to vastly reduce the computational requirement normally used in surrogate models based on the well-known Kriging approach~\cite{cressie1990origins} by using a Kriging Partial Least Square (KPLS) approach~\cite{bouhlel2016improving} and, secondly, to demonstrate the validity of using the phenotype distance vector on larger network sizes~\cite{pheno_dist_kernel}.

\begin{table}[ht!]

\caption{Details of the classification data sets used in our research as well as preliminary results for an initial surrogate model of 100 networks. A '-' indicates the experiment did not complete within 48 hrs. }
\centering
\resizebox{0.48\textwidth}{!}{ 
\begin{tabular}{llrrrrr}
\hline
 Dataset & Brief description & Pheno. dist.  & KPLS & Kriging \\
 & &  vector & & \\
 \hline
 
Iris  & 3 classes (Type of Iris plant)    & 336    & 00:00:25 & 01:52:38

      \\

Yeast & 2 classes (Protein sequence)      &  1338 & 00:02:24	& -
 \\

Ecoli & 8 classes (Localization site)  & 2016  & 00:01:09 &	-
   \\

Abalone & 2 classes (Number of rings)    & 6264 & 00:10:38	& -
\\

\hline
\end{tabular}
}
\vspace{-0.4 cm}
\label{tab:datasets}
\end{table}

\section{Methodology}
A major impediment in the use of SAEAs in neuroevolution is in the computational power still required when using well-known SAEAs approaches, such as the use of the Kriging method. To correctly address this, we
use and store ANNs’ behaviours in vectors, employ a population-based surrogate model strategy~\cite{jin2011surrogate}, and  apply Kriging Partial Least Squares  (KPLS) using the stored vectors~\cite{pheno_dist_kernel}. For full details of the Kriging and KPLS method see ~\cite{bouhlel2016improving}. More specifically, we use the \textit{behaviours} of the ANNs encoded in our EAs candidates solutions. These behaviours refer to the outputs obtained when these ANNs are executed over an input data set. A similar notion of behaviour (semantics) of individuals~\cite{Uy2011} has been widely adopted by the Genetic Programming community~\cite{9308386, 10.1145/3520304.3534073, galvan2022semantics, Uy2011, stapleton2021semantic, galvan2020promoting}. 
In our work, we design a surrogate-assisted evolutionary algorithm to approximate the fitness function of our evolutionary computation system. This is then used in a subset of the population at each generation. To this end, we use a population-based surrogate model strategy~\cite{jin2011surrogate}. 

It is well-known that stochastic gradient descent (SGD) is the most effective method to train ANNs~\cite{Galvn2021NeuroevolutionID}. To account for this key element, we adopt dCGPANN~\cite{izzo2017differentiable, martens2019neural}, a variation of Cartesian Genetic Programming (CGP)~\cite{Miller2011}. dCGPANN allows us to differentiate weights, and so, we can use SGD instead of evolving the ANNs’ weights. Each dCGPANN individual representing an ANN is then trained to a limited number of epochs. This results in using the behaviours of the networks. That is, we generate a phenotypic distance vector, similar to the work by Stork et al.~\cite{pheno_dist_kernel}, where the vector contains the output of the nodes at the final layer for all data instances. We use the KPLS method, by employing the SMT toolbox~\cite{SMT2019}, to estimate the fitness values of the subset of the population by feeding these vectors. 

%The rest of the population is evaluated as usual by using the fitness function which is the mean squared error of the training data.

\section{Experimental setup and results}

To test the efficiency of our methods, we use data sets from the UCI machine learning repository~\cite{Dua:2019}.  The third column, in Table~\ref{tab:datasets}, denotes the phenotype vector length. Larger phenotype vectors result in higher dimensional data for the surrogate model to train on. All experiments were run on Intel Xeon Gold 6148. Table~\ref{tab:datasets} summarises the preliminary results, showing the time taken to construct an initial population of 100 fully trained networks that are then used to construct the surrogate model. A dash symbol (-) indicates the experiment did not complete within 48 hours. We can see that based on the last two columns it is infeasible to use the Kriging approach for these datasets while for the KPLS approach, the times would be more than sufficient.

\section{Conclusions}
Surrogate-assisted evolutionary algorithms (SAEAs) are an active area of research, with limited research carried out in the area of neuroevolution, a method highly popular in the autonomous configuration of ANNs’ hyperparameters, weights and architectures. The use of common SAEAs techniques such as Kriging cannot be used in neuroevolution due to the high dimensional data normally used in neuroevolution. We have taken the initial steps to show that this can be addressed by adopting a correct representation and using the Kriging Partial Least Square method. Further studies are required to generalise our findings.

\section{Acknowledgments}

This publication has emanated from research conducted with the financial support of Science Foundation Ireland under Grant number 18/CRT/6049. For the purpose of Open Access, the author has applied a CC BY public copyright licence to any Author Accepted Manuscript version arising from this submission.

\bibliographystyle{abbrv}
\bibliography{ref}

\end{document}